\documentclass{ifacconf}

\usepackage{graphicx}
\usepackage{natbib} 

\usepackage{amsmath}
\DeclareMathOperator{\dd}{\mathrm{d}}
\DeclareMathOperator*{\argmax}{arg\,max}
\usepackage{bm}
\usepackage{nicefrac} 

\usepackage{xcolor}

\usepackage{url}
\begin{document}
\begin{frontmatter}

\title{Experimental Design for Missing Physics}

\author[First,Second,Third]{Arno Strouwen} 
\author[Second,Fourth]{Sebastian Micluța-Câmpeanu}

\address[First]{Strouwen Statistics (e-mail: contact@arnostrouwen.com)}
\address[Second]{JuliaHub, MA 02111-1929 USA}
\address[Third]{Biosystems Department, 
   KULeuven, Leuven, Belgium}
\address[Fourth]{Faculty of Physics, University of Bucharest, Bucharest, Romania}
\begin{abstract} 
For most process systems, knowledge of the model structure is incomplete.
This missing physics must then be learned from experimental data.
Recently, a combination of universal differential equations and symbolic regression has become a popular tool to discover these missing physics.
Universal differential equations employ neural networks to represent missing parts of the model structure,
and symbolic regression aims to make these neural networks interpretable.
These machine learning techniques require high-quality data to successfully recover the true model structure.
To gather such informative data, a sequential experimental design technique is developed which is based on optimally discriminating between the plausible model structures suggested by symbolic regression. This technique is then applied to discovering the missing physics of a bioreactor.
\end{abstract}

\begin{keyword}
Optimal Experimental Design, Missing Physics, Neural Networks, Universal Differential Equation, Symbolic Regression, Model Discrimination.
\end{keyword}

\end{frontmatter}
\renewcommand{\thefootnote}{}\footnotetext{\textcopyright~2025 the authors. This work has been accepted to IFAC for publication under a Creative Commons Licence CC-BY-NC-ND.}\renewcommand{\thefootnote}{\arabic{footnote}}

\section{Introduction}
Model-based approaches are commonly used in the analysis, control, and optimization of process (bio)systems.
These models rely on knowledge of physical, chemical, and biological laws,
such as conservation laws, transport phenomena, and reaction kinetics,
which are usually described by a system of nonlinear differential equations.

Often, our knowledge of the laws acting on the system is incomplete.
These gaps in our knowledge are also referred to as missing physics.
Experimental data can be used to fill in such missing physics \citep{harlim}.
Recently, Universal Differential Equations (UDE) were proposed to learn the missing parts of the structure \citep{rackauckas1}.
Universal Differential Equations use neural networks to represent the terms of the model for which the underlying structure is unknown \citep{dandekar}.

Because the opaque nature of neural networks is often not desirable in a scientific computing setting,
UDE based techniques are often combined with interpretable machine learning techniques,
such as sparse regression \citep{kaiser} or symbolic regression \citep{koza}. These techniques post-process the neural network into a human-readable model structure.

Universal differential equations are quickly gaining in popularity, with multiple applications in physics \citep{keithLearningOrbitalDynamics2021}, chemistry \citep{santanaEfficientHybridModeling2023} as well as biology \citep{philippsNonNegativeUniversalDifferential2024, rojas-camposLearningCOVID19Regional2023}.
Because neural networks are data-hungry \citep{van}, it is important that these applications gather highly informative data.
However, current model-based design of experiment (MbDoE) methodology focuses on parameter precision or discriminating between a finite number of possible model structures.
A review of both methods can be found in \citet{franceschini}.
When part of the model structure is entirely unknown, neither of these techniques can be directly applied.

In this paper, we propose an efficient data gathering technique for filling in missing physics with a universal differential equation, made interpretable with symbolic regression.
In particular, a sequential experimental design technique is developed, where an experiment is performed to discriminate between the plausible model structures suggested by symbolic regression. The new data is then used to retrain the UDE, which leads to a new set of plausible model structures by applying symbolic regression again.

This methodology is applied to a bioreactor, and is shown to perform better than a randomly controlled experiment.

\section{Missing Physics}
We illustrate the concept of missing physics with a well-mixed fed-batch bioreactor example. This reactor has a long history in the MbDoE literature \citep{versyck, telen, telen2, houska}.
The reactor has three dynamic states:
the substrate concentration, $C_s$,
the biomass concentration, $C_x$,
and the volume of the reactor, $V$.
The evolution in time of these states is governed by the following differential equations:
\begin{equation}\label{eq:bioreactor}
\begin{aligned}
\frac{dC_s}{dt} &= -\left(\frac{\mu(C_s)}{y_{x,s}} + m\right) C_x + \frac{Q_{in}(t)}{V}(C_{S,in} - C_s),\\
\frac{dC_x}{dt} &= \mu(C_s) C_x - \frac{Q_{in}(t)}{V}C_x,\\
\frac{dV}{dt} &= Q_{in}(t).
\end{aligned}
\end{equation}
In these equations, the specific growth rate, $\mu$, is an unknown function.
This function has a single input, $C_s$.
This unknown function must be determined from experimental data.
The true function that must be recovered is the Monod equation:
\begin{equation}
\mu(C_s) = \frac{\mu_{max}C_s}{K_s + C_s}.
\end{equation}
The experimental data we gather from a single bioreactor are $15$ hourly measurements of $C_s$, subject to measurement noise:
\begin{equation}
C_s^{\text{measured}}(t_k) = C_s(t_k) + \epsilon_{k},
\end{equation}
with the variance of the noise equal to $0.1^2$.

We do not perform only a single experiment,
and thus do not only gather a single time series for $C_s$.
Instead, we take a sequential approach to data gathering.
For each experiment, the volumetric feed rate, $Q_\text{in}(t)$,
will be optimized to gain as much information as possible about the missing physics, $\mu$.

The initial conditions, $C_s(t=0)$, $C_x(t=0)$ and $V(t=0)$, are assumed to be known constants,
as well as the substrate concentration in the feed, $C_{S,\text{in}}$,
the yield, $y_{x,s}$,
and the maintenance factor, $m$,
with numerical values taken from \citet{telen2}.
The missing physics, $\mu$, also further depends on two parameters,
the maximal specific growth rate, $\mu_\text{max}$,
and the half saturation constant, $K_s$,
with true values equal to $0.421$ $\mathrm{1}/\mathrm{h}$ and $4.39$ $\mathrm{g}/\mathrm{l}$, respectively.

More abstractly, we consider systems of the following form:
\begin{equation}\label{eq:system}
\begin{aligned}
	\frac{\dd \bm x}{\dd t} &= \bm f(t,\bm x,\bm \phi(\bm g(\bm x)),\bm u(t)), \qquad \text{with } \bm x(t=0) = \bm x_0;\\
	\bm y_k &= \bm h(\bm x(t_k)) + \bm \epsilon_k,
\end{aligned}
\end{equation}
where $t$ denotes the time ranging from $0$ to $t_e$,
the end time of the experiment.
The column vector $\bm y_k$ contains the measurements taken at a time point $t_k$,
with $k$ ranging from $1$ to $N$,
the number of measurement times.
The time between measurements is equally spread,
so that $t_k = \nicefrac{kt_e}{N}$.
A measurement at the end of the experiment is thus included,
but not at the start.
The measurements are subject to independent Gaussian noise.
More specifically, each $\bm \epsilon_k$ is identically and independently multivariate normally distributed with zero mean and covariance matrix $R$.
The measurements depend on the dynamic state column vector $\bm x(t)$ through the measurement function $\bm h$.
This function $\bm h$ is useful, for example, when not all states are measured, such as in the bioreactor example.
The states $\bm x(t)$ have to be calculated from the system of ordinary differential equations $\bm f$,
with initial conditions $\bm x_0$.

This system $\bm f$ depends on the output of the function $\bm \phi$. This function represents the missing physics, i.e. the parts of the model structure which are unknown. The input of the function $\bm \phi$ is not directly the state $\bm x$, but instead the output of another known preprocessing function $\bm g$, which in turn has the state $\bm x$ as an input: $\bm \phi(\bm g(\bm x))$. The function $\bm g$ is useful, for example, when we know the missing physics only depends on a subset of the states, such as in the bioreactor example.

The system $\bm f$ also depends on the controllable input column vector $\bm u(t)$,
which we will optimize to determine $\bm \phi$ as precisely as possible.
Finding the optimal controls is an infinite dimensional optimization problem.
To reduce the complexity of this problem to a non-linear optimization one,
we restrict $\bm u(t)$ to piecewise constant functions, which are allowed to jump whenever a measurement is gathered:
\begin{equation}
\bm u(t) = \sum_{k=1}^N \bm u_k \text{rect}\left(\frac{t-t_{k-1}-0.5}{t_k-t_{k-1}}\right), \quad \bm u_\text{min} \leq \bm u_k \leq \bm u_\text{max},
\end{equation}
where rect is the rectangular function \citep{tang}.
The coefficients $\bm u_k$ must then be optimized to give as much information as possible, and are constrained between a minimal and maximal control value of $\bm u_\text{min}$ and $\bm u_\text{max}$.
For the bioreactor example, these extrema are again taken from \citep{telen}.
\section{Universal Differential Equation}
Component based UDE replace the unknown function $\bm \phi$ in (\ref{eq:system}) with a neural network.
\begin{equation}\label{eq:UDE}
	\frac{\dd \bm x}{\dd t} = \bm f(t,\bm x,\text{NN}(\bm g(\bm x), \bm \theta),\bm u(t)).
\end{equation}
In this equation $\text{NN}(\bm g(\bm x), \bm \theta)$ is the neural network.
Similar to $\bm \phi(\bm g(x))$, the input of the neural network is $\bm g(x)$,
but the network is also dependent on the parameters $\bm \theta$,
which must be learned from the experimental data.

Concretely applied to the bioreactor example, this becomes:
\begin{equation}\label{eq:bioreactorUDE}
\begin{aligned}
\frac{dC_s}{dt} &= -\left(\frac{\includegraphics[width=1cm]{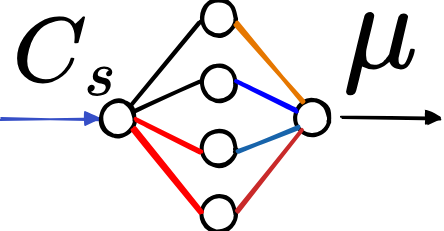}}{y_{x,s}} + m\right) C_x + \frac{Q_{in}(t)}{V}(C_{S,in} - C_s),\\
\frac{dC_x}{dt} &= \includegraphics[width=1cm]{nn.png}\ C_x - \frac{Q_{in}(t)}{V}C_x,\\
\frac{dV}{dt} &= Q_{in}(t).
\end{aligned}
\end{equation}

\begin{figure*}
\begin{center}
\includegraphics[width=12cm]{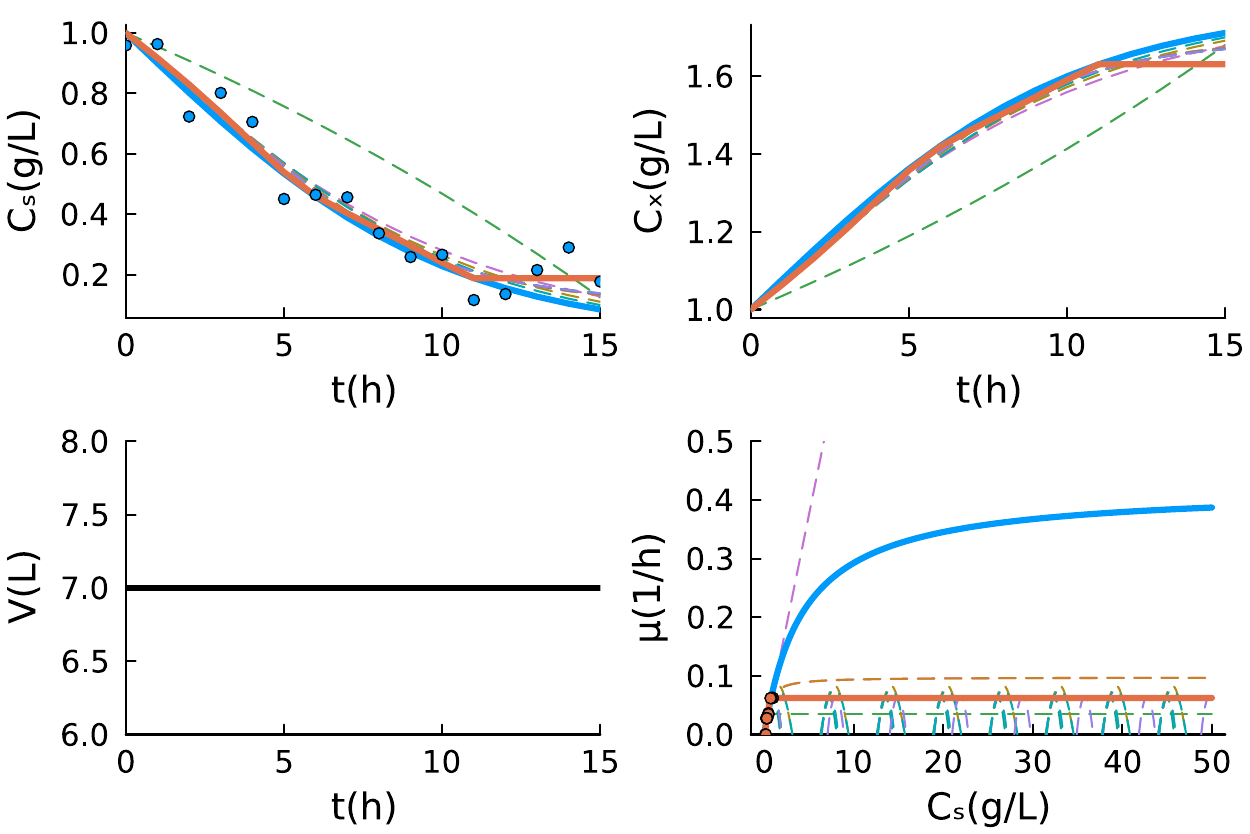}    
\caption{First experiment. The three states of the bioreactor, and the missing physics $\mu$.
The blue solid lines corresponds to the true system,
while the blue dots correspond to measured values.
The orange solid lines correspond to predictions made by the UDE,
while the orange dots correspond to the predicted values at measured $C_s$.
The dashed lines correspond to predictions made by the plausible model structures.
For the state $V$, all these lines coincide, and have been replaced by a black line.}
\label{fig:experiment1}
\end{center}
\end{figure*}

\section{Symbolic Regression}

In order to gain more physical insight into the fitted neural network, we turn to symbolic regression.
Symbolic regression is an algorithmic way of searching the space of mathematical expressions to find the model that best fits a given dataset, while balancing accuracy and simplicity. Specifically, tree data structures are used to represent mathematical expressions, and genetic algorithms are used to search for accurate trees, while keeping the trees as small as possible.

For the input, we use the values of the states predicted by the universal differential equation at the measurement times, after they have passed through the preprocessing function, $\bm g(\bm x(t_k))$,
while for the output, we use the predicted values of the trained neural network at the measurement times, 
$NN(\bm g(\bm x(t_k), \bm \theta)$.

This will result in a number of plausible symbolic expressions that agree well with the neural network output at the measurement times.
We then rank the top $M$ candidates using the default option in SymbolicRegression.jl, which is based on a measure of the precision and complexity of the expressions, where the complexity is defined as ``the number of nodes in an expression tree, regardless of each node's content'' \citep{cranmerInterpretableMachineLearning2023}.
\section{Efficient Data Gathering for Missing Physics}
We now wish to discriminate between the plausible model structures suggested by symbolic regression. We will do this by creating a variant of T-optimal designs \citep{ucinski}. T-optimal designs are model discrimination designs, where design points are sought which maximize the difference between the predicted output of a model thought to be correct (T for true) and the predicted output of some other plausible alternative model structures. It should then be easy to discern from the gathered data if the “true” model is really correct after all.

In our situation, we do not have a model structure which can serve as the ground truth. We will instead work with all pairwise distances between the plausible model structures suggested by symbolic regression:
\begin{equation}\label{eq:optimal}
\argmax_{\bm u} \frac{2!(M-2)!}{M!}\sum_{i=1}^M \sum_{j=i+1}^M \max_{t_k} (h(\bm x_i(t_k)) - h(\bm x_j(t_k)))^2.
\end{equation}
In this equation, $\bm x_i$ denotes the predicted states for the i'th plausible model structure.
The distance between two model structures is scored by the maximal squared difference between the two structures at the measurement times. The criterion then calculates the average distance between all model structures. Collecting measurements where the plausible model structures differ greatly in predictions, will cause at least some of the model structures to become unlikely, and thus cause new model structures to enter the top $M$ plausible model structures.
We can then continue by constructing a next experiment using the new top $M$ plausible model structures.

\begin{figure}[]
\begin{center}
\includegraphics[width=8cm]{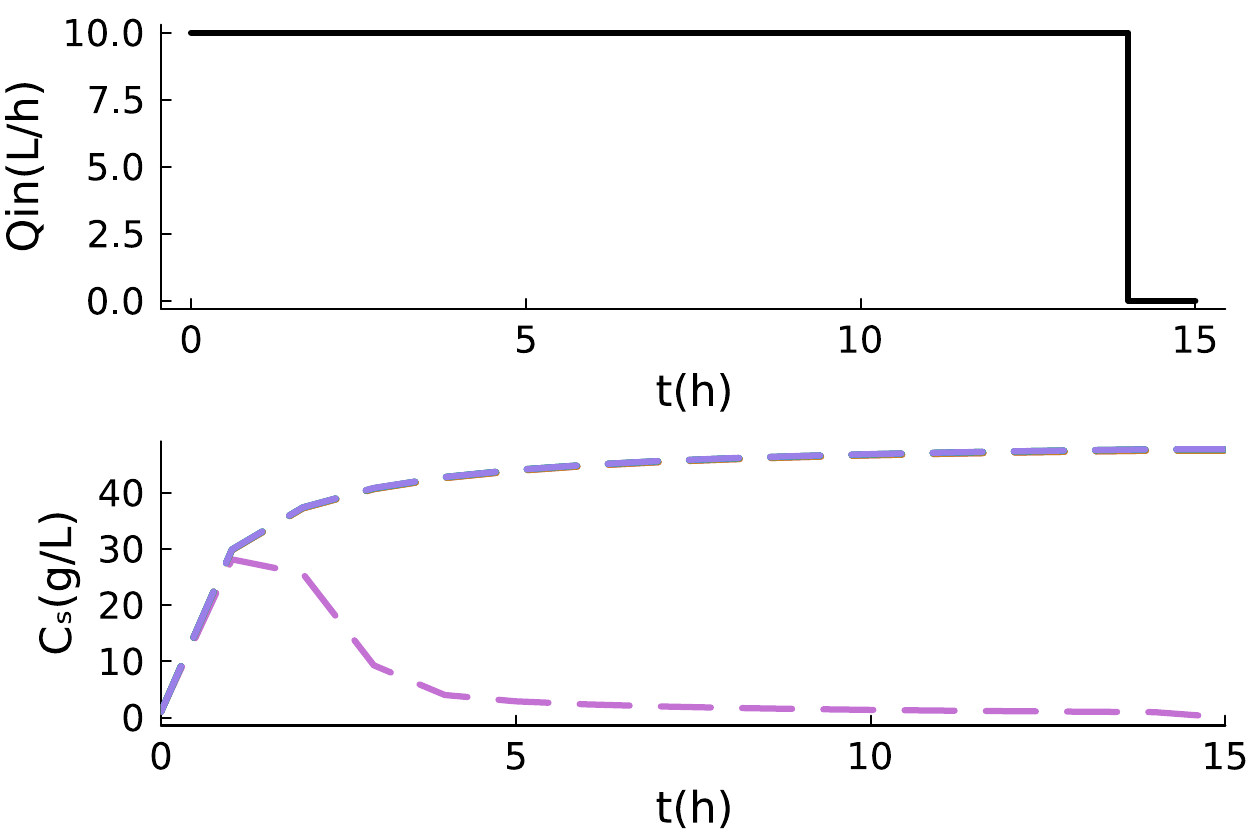}    
\caption{Top: Optimal control for the second experiment.
Bottom: predictions made by the plausible model structures.} 
\label{fig:control1}
\end{center}
\end{figure}

\section{Results}
To start the sequential design process, we gather some initial data, using the zero function as the control signal. Fig. \ref{fig:experiment1} shows the data analysis for this initial experiment. The UDE generally predicts the true states well, except during the last three hours.
This is because the realization of the measurement noise was highly positive during these three measurements.

At the measured values of $C_s$, the missing physics, $\mu$, is also approximated well by the UDE.
However, at larger values of $C_s$, where we do not have any measurements, the UDE does not fit the true function well, which is not surprising since neural networks are not able to extrapolate outside the region where data has been gathered.

The plausible model structures, suggested by symbolic regression, also predict the states well. There is one exception to this, the green dashed line corresponds to $\mu$ being a constant, which is a too simplistic structure to predict the states well.
Similar to the UDE, the plausible model structures also fit $\mu$ well in the low $C_s$ region, but not outside this region. One group of the structures predicts that $\mu$ keeps increasing as $C_s$ becomes larger, while another group predicts that $\mu$ stays below $0.1$ $1/\mathrm{h}$.
We now design a second experiment to start discriminating between these model structures.

The optimal control for the second experiment, as shown in Fig. \ref{fig:control1},
prefers to use the highest allowed value for the controls almost during the entire experiment. This control action allows us to easily discriminate between the two aforementioned groups, because the predicted $C_s$ differs greatly for these two groups.

\begin{figure*}
\begin{center}
\includegraphics[width=12cm]{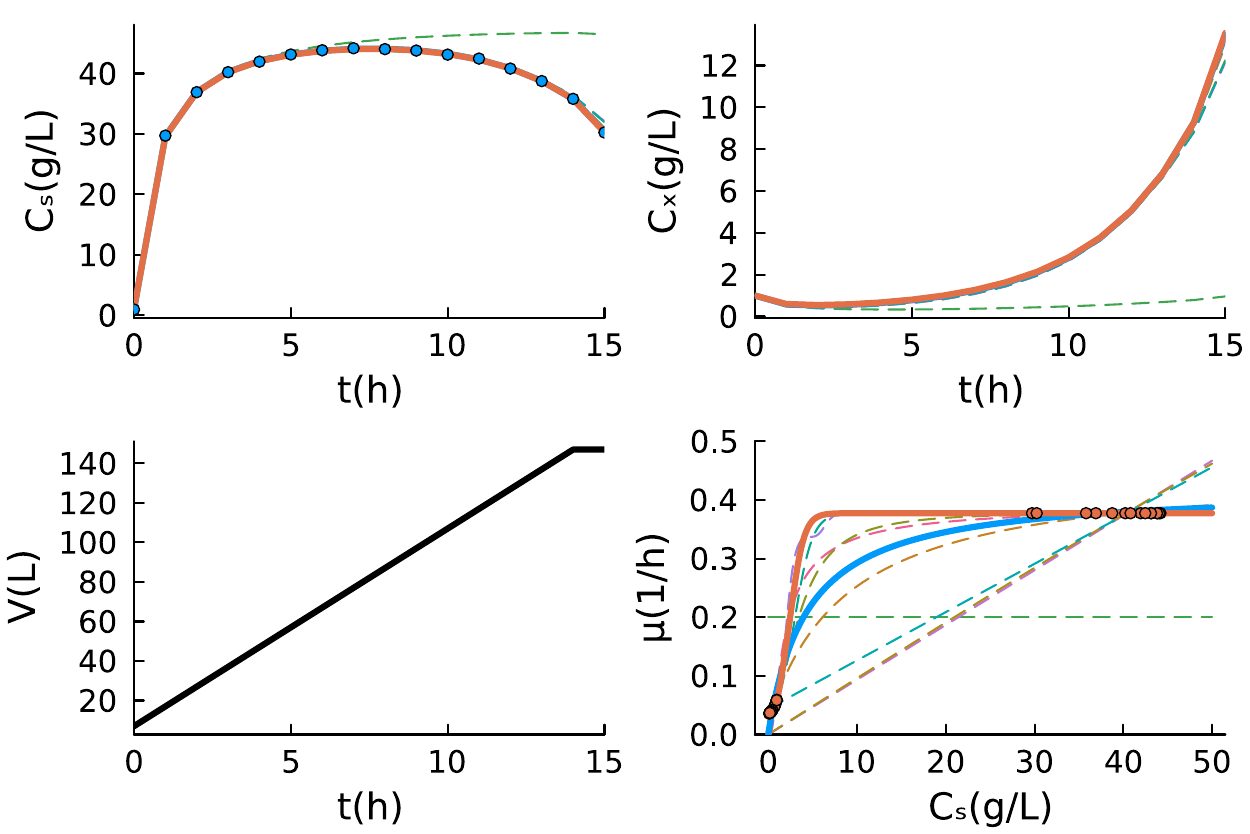}    
\caption{Second experiment. The three states of the bioreactor, and the missing physics $\mu$.
The blue solid lines corresponds to the true system,
while the blue dots correspond to measured values.
The orange solid lines correspond to predictions made by the UDE,
while the orange dots correspond to the predicted values at measured $C_s$.
The orange dots not only represent predictions for the second experiment, but also the first.
The dashed lines correspond to predictions made by the plausible model structures.
For the state $V$, all these lines coincide, and have been replaced by a black line.}
\label{fig:experiment2}
\end{center}
\end{figure*}

Fig. \ref{fig:experiment2} shows the data analysis corresponding to this second experiment.
Both the UDE and most of the plausible model structures predict the states well, with the same exception of the constant function as in the previous experiment.

The UDE and the plausible model structures (except the constant one) also approximate the missing physics $\mu$ well in the region where we have gathered data. This means in the regions of low substrate concentration, with data coming primarily from the first experiment, and high substrate concentration, coming from the second experiment. However, we do not have any measurements at substrate concentrations between these two groups. This causes there to be substantial disagreement between the plausible model structures in the medium substrate concentration range.
One of the plausible model structures is the Monod equation, with reasonably accurate parameter values: $\nicefrac{C_s}{(C_s - (-5.4499))}*0.42887$. Symbolic regression sometimes finds the true model structure in a somewhat unusual form, like with a double negative sign. This is because symbolic regression considers addition and subtraction to have the same complexity, as well as positive and negative numbers.

\begin{figure}
\begin{center}
\includegraphics[width=8cm]{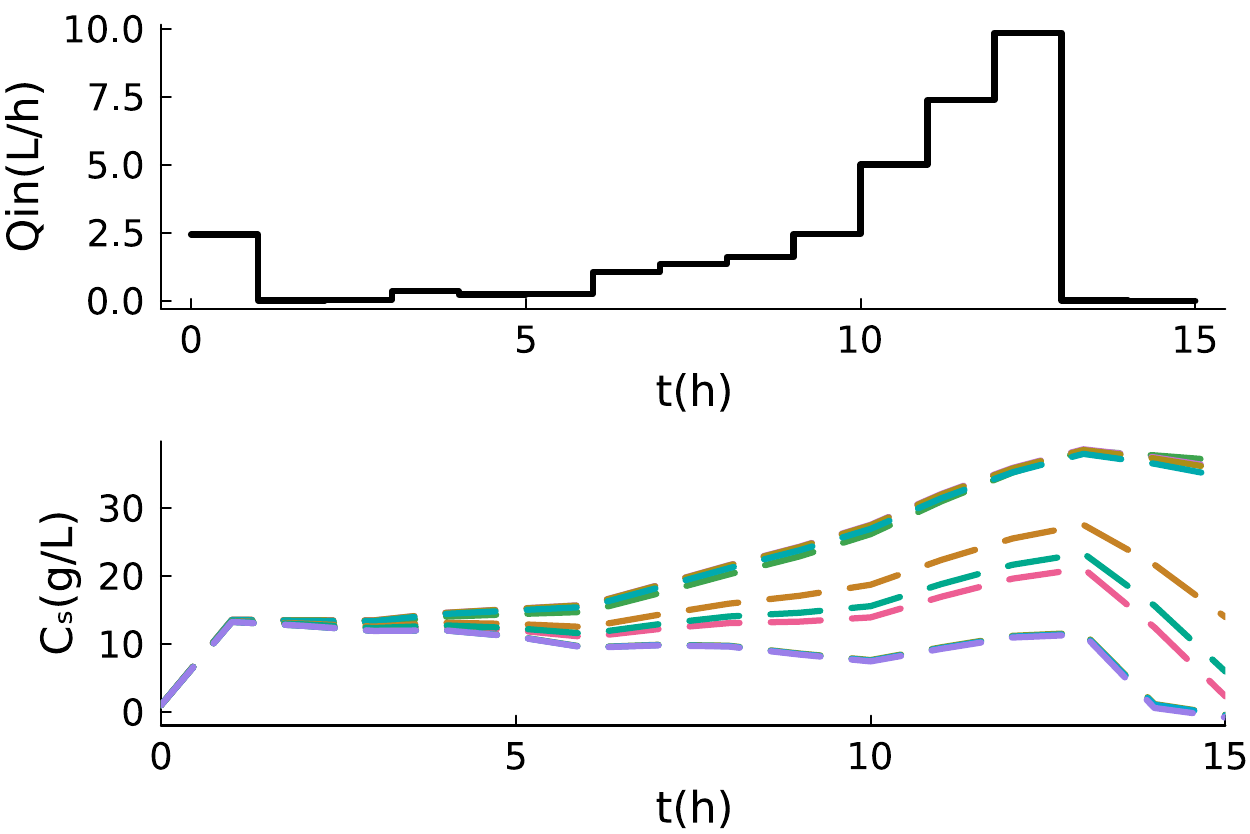}    
\caption{Top: Optimal control for the third experiment.
Bottom: predictions made by the plausible model structures.} 
\label{fig:control2}
\end{center}
\end{figure}

The optimal design algorithm is also aware of this uncertainty at the medium concentration range,
and aims to remedy this in the next experiment, as can be seen on Fig. \ref{fig:control2}.
Using the first control action, the bioreactor substrate concentration gets pumped from a low substrate concentration level to a medium level. At this level, there is substantial disagreement between the plausible model structures, leading to substantial disagreement in predicted substrate concentrations.
To keep the reactor at the medium substrate concentration range, while the biomass concentration increases rapidly,
an increasing amount of substrate has to be pumped into the reactor every hour.
This explains the staircase with increasing step heights form of the control function.
After the staircase reaches the maximal control value, a zero control is used.
Some model structures decrease more rapidly in substrate concentration than others.

\begin{figure}
\begin{center}
\includegraphics[width=8.4cm]{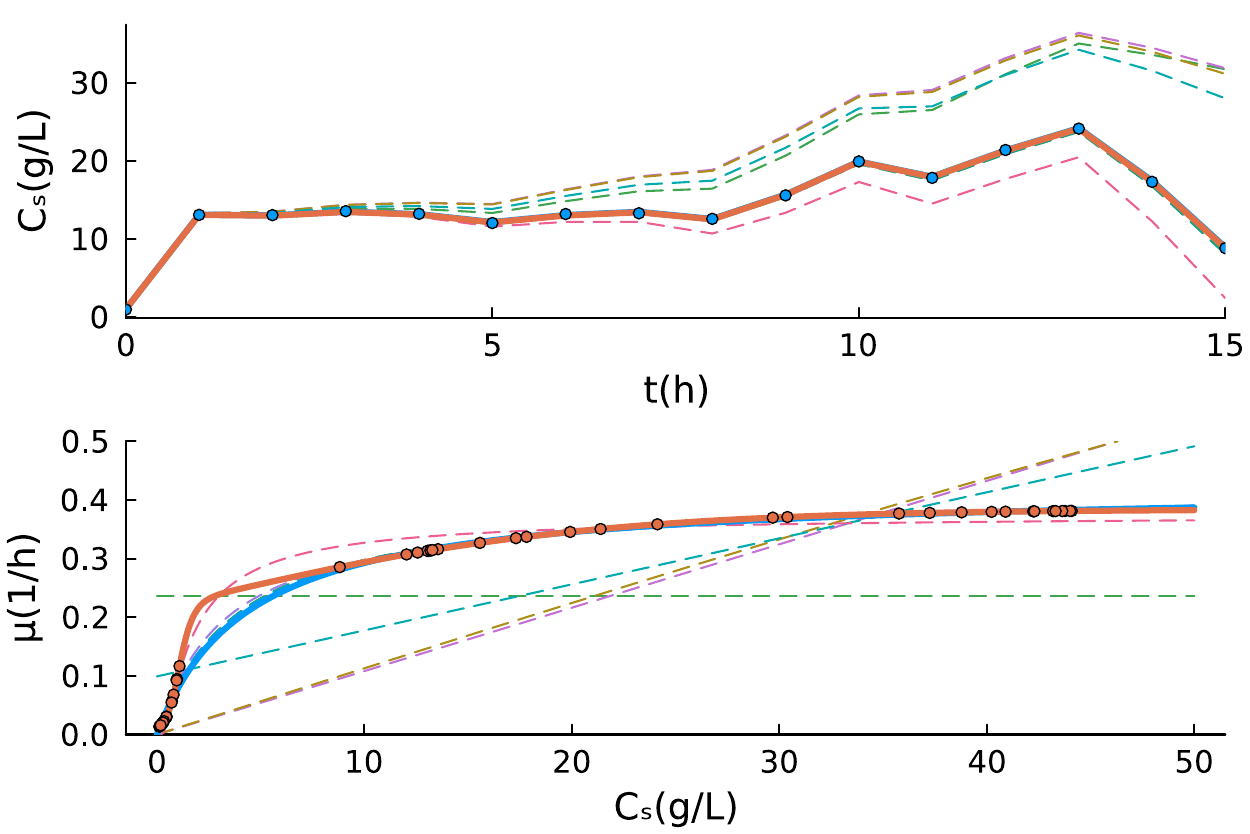}    
\caption{Selected output of the third experiment: The state $C_s$, and the missing physics $\mu$.
The blue solid lines corresponds to the true system,
while the blue dots correspond to measured values.
The orange solid lines correspond to predictions made by the UDE,
while the orange dots correspond to the predicted values at measured $C_s$.
The orange dots not only represent predictions for the third experiment, but also the first and second.
The dashed lines correspond to predictions made by the plausible model structures.}
\label{fig:experiment3}
\end{center}
\end{figure}

After the tree optimal experiments have been performed, the Monod kinetics score the highest using the default way of scoring model structures by SymbolicRegression.jl, as shown in Table \ref{tb:HOF}. This suggests that the Monod kinetics is a highly plausible model structure \citep{cranmerInterpretableMachineLearning2023}. These models structures are depicted in Fig. \ref{fig:experiment3}. Model structures less complex than the Monod kinetics underfit.

We also performed 5 random experiments, each consisting of 3 time series,
where the controls $\bm u_k$ were drawn from a uniform distribution, with minimum $\bm u_\text{min}$ and maximum $\bm u_\text{max}$.
The data coming from these experiments was analyzed the same way as the optimal experiments.
However, the Monod equation was not recovered in any of the 5 random experiments.
This shows that there is an information gain by using optimal experimental design techniques.
\begin{table}[h]
\begin{center}
\caption{Symbolic regression hall of fame after three experiments. Monod kinetics has the largest score.}
\label{tb:HOF}
\begin{tabular}{ll}
Score & Equation  \\\hline
3.604e+01  &0.23468\\
4.571e-01  &$C_s$ * 0.010807\\
5.788e-02  &sin($C_s$ * 0.011293)\\
7.005e-01  &$C_s$ / ($C_s$ - -50.467)\\
2.787e+00  &exp(-1.5505 / $C_s$) * 0.37874\\\hline
3.076e+00  &$C_s$ / (($C_s$ - -4.631) / 0.42347)\\\hline
1.140e-02  &sin($C_s$ / (($C_s$ - -4.8269) / 0.43601))\\
4.483e-01  &($C_s$ / (($C_s$ - -4.2621) / 0.42527)) - 0.0059018\\
2.075e-01  &($C_s$ -(0.014069/$C_s$))/(($C_s$+4.4614)/0.42113)\\
1.474e-04  &{\small ($C_s$-sin(0.014091/$C_s$))/(($C_s$+4.4613)/0.42113)}\\
\end{tabular}
\end{center}
\end{table}
\section{Computational details}
\label{sec:CD}
All simulations are implemented using the Julia programming language \citep{bezanson},
in particular using the ModelingToolkit.jl framework \citep{ma}.
All differential equations are solved using DifferentialEquations.jl \citep{rackauckas2},
specifically the Rodas5P solver \citep{steinebach}.
The discontinuities in the piecewise-constant controls are implemented using the discrete-time callback functionality of DifferentialEquations.jl.
All optimization problems are solved using Optimization.jl \citep{vaibhav}.
The neural network fitting uses L-BFGS \citep{liu}, on the L2-loss function.
The controls are optimized for $100$ $\mathrm{s}$ using the adaptive\_de\_rand\_1\_bin\_radiuslimited method \citep{feoktistov}, from BlackBoxOptim.jl \citep{feldt}.
If a differential equation fails to solve, inside an optimization objective, then the worst possible value for that objective is returned.
A neural network with two hidden layers with 5 units each is used.
The hyperbolic tangent is used as the activation function for the hidden layers,
while the sigmoid function is used for the output layer.
The network is implemented in Lux.jl \citep{pal}.
Symbolic regression is implemented using SymbolicRegression.jl \citep{cranmerInterpretableMachineLearning2023}.
Symbolic regression is allowed to run for 1000 iterations, with parallelism disabled,
and deterministic mode enabled. The operators in the symbolic search space consist of
the exponential, sine and cosine functions, as well as the addition, subtraction, multiplication and division functions.
The top 10 model structures are tracked, $M=10$. Symbolic regression sometimes suggests the same model structures in different forms, e.g. $\nicefrac{1}{\nicefrac{1}{x}}$ and $x$.
To protect against such duplicates, we do not track structures of higher complexity, which have the same L2-loss as a lower complexity model up to 5 significant digits.

Any hyperparameter not mentioned here keeps the default value provided by aforementioned software packages.

The source code accompanying this paper can be found on \url{https://github.com/arno-papers/DYCOPS2025}.
\section{Discussion}
In (\ref{eq:system}) the only unknown part of the system is the function $\bm \phi$. In many realistic scenarios, there will not only be missing physics, but also parameters which much be tuned to the experimental data. In theory, these cases can be covered by our methodology, by considering the parameters as constant functions, which must be learned. However, specialized experimental design techniques for precisely estimating such parameters are well known \citep{franceschini}. The criteria for discovering missing physics and estimating parameters precisely could then be combined using multi-objective model-based experimental design \citep{telen}.

In our examples, the only design variable to be optimized was the control $\bm u(t)$. In the model-based experimental design literature, it is also common to optimize other aspects of the design, such as measurement times, duration of the experiment and initial conditions \citep{galvanin}. Incorporating these other aspects, would be a straightforward modification of our methodology.

The missing physics of the bioreactor, $\bm \mu$, is a function with a single input and output.
This made it easy for us to visualize the added value of the optimal design. However, the methodology proposed in this paper also works for missing physics with multiple inputs and outputs. A similar comment holds for the controls.

The different steps of our methodology: training the neural network, performing symbolic regression on the neural network, and discriminating between the suggested model structures all have their own associated hyperparameters, as detailed in section \ref{sec:CD}. Currently, these hyperparameters are fixed throughout the entire experiment. A true online experimental design methodology would require that these hyperparameters are also tuned automatically. We consider this automatic tuning of hyperparameters to be one of the most interesting directions for future research.
\bibliography{ifacconf}

\begin{thebibliography}{4}
\providecommand{\natexlab}[1]{#1}
\providecommand{\url}[1]{\texttt{#1}}
\providecommand{\urlprefix}{URL }
\expandafter\ifx\csname urlstyle\endcsname\relax
  \providecommand{\doi}[1]{doi:\discretionary{}{}{}#1}\else
  \providecommand{\doi}{doi:\discretionary{}{}{}\begingroup
  \urlstyle{rm}\Url}\fi

\bibitem[{Able(1956)}]{Abl:56}
Able, B. (1956).
\newblock Nucleic acid content of microscope.
\newblock \emph{Nature}, 135, 7--9.

\bibitem[{Able et~al.(1954)Able, Tagg, and Rush}]{AbTaRu:54}
Able, B., Tagg, R., and Rush, M. (1954).
\newblock Enzyme-catalyzed cellular transanimations.
\newblock In A.~Round (ed.), \emph{Advances in Enzymology}, volume~2, 125--247.
  Academic Press, New York, 3rd edition.

\bibitem[{Keohane(1958)}]{Keo:58}
Keohane, R. (1958).
\newblock \emph{Power and Interdependence: World Politics in Transitions}.
\newblock Little, Brown \& Co., Boston.

\bibitem[{Powers(1985)}]{Pow:85}
Powers, T. (1985).
\newblock Is there a way out?
\newblock \emph{Harpers}, 35--47.

\end{thebibliography}


\begin{thebibliography}{28}
\providecommand{\natexlab}[1]{#1}
\providecommand{\url}[1]{\texttt{#1}}
\providecommand{\urlprefix}{URL }
\expandafter\ifx\csname urlstyle\endcsname\relax
  \providecommand{\doi}[1]{doi:\discretionary{}{}{}#1}\else
  \providecommand{\doi}{doi:\discretionary{}{}{}\begingroup
  \urlstyle{rm}\Url}\fi

\bibitem[{Bezanson et~al.(2017)Bezanson, Edelman, Karpinski, and
  Shah}]{bezanson}
Bezanson, J., Edelman, A., Karpinski, S., and Shah, V.B. (2017).
\newblock Julia: A fresh approach to numerical computing.
\newblock \emph{SIAM review}, 59(1), 65--98.

\bibitem[{Cranmer(2023)}]{cranmerInterpretableMachineLearning2023}
Cranmer, M. (2023).
\newblock Interpretable {{Machine Learning}} for {{Science}} with {{PySR}} and
  {{SymbolicRegression}}.jl.
\newblock \doi{10.48550/arXiv.2305.01582}.

\bibitem[{Dandekar et~al.(2020)Dandekar, Chung, Dixit, Tarek, Garcia-Valadez,
  Vemula, and Rackauckas}]{dandekar}
Dandekar, R., Chung, K., Dixit, V., Tarek, M., Garcia-Valadez, A., Vemula,
  K.V., and Rackauckas, C. (2020).
\newblock Bayesian neural ordinary differential equations.
\newblock \emph{arXiv preprint arXiv:2012.07244}.

\bibitem[{Dixit and Rackauckas(2023)}]{vaibhav}
Dixit, V.K. and Rackauckas, C. (2023).
\newblock Optimization.jl: A unified optimization package.
\newblock \doi{10.5281/zenodo.7738525}.
\newblock \urlprefix\url{https://doi.org/10.5281/zenodo.7738525}.

\bibitem[{Feldt and Stukalov(2018)}]{feldt}
Feldt, R. and Stukalov, A. (2018).
\newblock Blackboxoptim. jl.
\newblock \emph{See https://github. com/robertfeldt/BlackBoxOptim. jl}.

\bibitem[{Feoktistov(2006)}]{feoktistov}
Feoktistov, V. (2006).
\newblock \emph{Differential evolution}.
\newblock Springer.

\bibitem[{Franceschini and Macchietto(2008)}]{franceschini}
Franceschini, G. and Macchietto, S. (2008).
\newblock Model-based design of experiments for parameter precision: State of
  the art.
\newblock \emph{Chemical Engineering Science}, 63(19), 4846--4872.

\bibitem[{Galvanin et~al.(2011)Galvanin, Boschiero, Barolo, and
  Bezzo}]{galvanin}
Galvanin, F., Boschiero, A., Barolo, M., and Bezzo, F. (2011).
\newblock Model-based design of experiments in the presence of continuous
  measurement systems.
\newblock \emph{Industrial \& Engineering Chemistry Research}, 50(4),
  2167--2175.

\bibitem[{Harlim et~al.(2021)Harlim, Jiang, Liang, and Yang}]{harlim}
Harlim, J., Jiang, S.W., Liang, S., and Yang, H. (2021).
\newblock Machine learning for prediction with missing dynamics.
\newblock \emph{Journal of Computational Physics}, 428, 109922.

\bibitem[{Houska et~al.(2015)Houska, Telen, Logist, Diehl, and
  Van~Impe}]{houska}
Houska, B., Telen, D., Logist, F., Diehl, M., and Van~Impe, J.F. (2015).
\newblock An economic objective for the optimal experiment design of nonlinear
  dynamic processes.
\newblock \emph{Automatica}, 51, 98--103.

\bibitem[{Kaiser et~al.(2018)Kaiser, Kutz, and Brunton}]{kaiser}
Kaiser, E., Kutz, J.N., and Brunton, S.L. (2018).
\newblock Sparse identification of nonlinear dynamics for model predictive
  control in the low-data limit.
\newblock \emph{Proceedings of the Royal Society A}, 474(2219), 20180335.

\bibitem[{Keith et~al.(2021)Keith, Khadse, and
  Field}]{keithLearningOrbitalDynamics2021}
Keith, B., Khadse, A., and Field, S.E. (2021).
\newblock Learning orbital dynamics of binary black hole systems from
  gravitational wave measurements.
\newblock \emph{Physical Review Research}, 3(4), 043101.
\newblock \doi{10.1103/PhysRevResearch.3.043101}.

\bibitem[{Koza(1994)}]{koza}
Koza, J.R. (1994).
\newblock Genetic programming as a means for programming computers by natural
  selection.
\newblock \emph{Statistics and computing}, 4, 87--112.

\bibitem[{Liu and Nocedal(1989)}]{liu}
Liu, D.C. and Nocedal, J. (1989).
\newblock On the limited memory bfgs method for large scale optimization.
\newblock \emph{Mathematical programming}, 45(1), 503--528.

\bibitem[{Ma et~al.(2021)Ma, Gowda, Anantharaman, Laughman, Shah, and
  Rackauckas}]{ma}
Ma, Y., Gowda, S., Anantharaman, R., Laughman, C., Shah, V., and Rackauckas, C.
  (2021).
\newblock Modelingtoolkit: A composable graph transformation system for
  equation-based modeling.
\newblock \emph{arXiv preprint arXiv:2103.05244}.

\bibitem[{Pal(2023)}]{pal}
Pal, A. (2023).
\newblock {On Efficient Training \& Inference of Neural Differential
  Equations}.

\bibitem[{Philipps et~al.(2024)Philipps, K{\"o}rner, Vanhoefer, Pathirana, and
  Hasenauer}]{philippsNonNegativeUniversalDifferential2024}
Philipps, M., K{\"o}rner, A., Vanhoefer, J., Pathirana, D., and Hasenauer, J.
  (2024).
\newblock Non-{{Negative Universal Differential Equations With Applications}}
  in {{Systems Biology}}.
\newblock \doi{10.48550/ARXIV.2406.14246}.

\bibitem[{Rackauckas et~al.(2020)Rackauckas, Ma, Martensen, Warner, Zubov,
  Supekar, Skinner, Ramadhan, and Edelman}]{rackauckas1}
Rackauckas, C., Ma, Y., Martensen, J., Warner, C., Zubov, K., Supekar, R.,
  Skinner, D., Ramadhan, A., and Edelman, A. (2020).
\newblock Universal differential equations for scientific machine learning.
\newblock \emph{arXiv preprint arXiv:2001.04385}.

\bibitem[{Rackauckas and Nie(2017)}]{rackauckas2}
Rackauckas, C. and Nie, Q. (2017).
\newblock Differentialequations. jl--a performant and feature-rich ecosystem
  for solving differential equations in julia.
\newblock \emph{Journal of Open Research Software}, 5(1).

\bibitem[{{Rojas-Campos} et~al.(2023){Rojas-Campos}, Stelz, and
  Nieters}]{rojas-camposLearningCOVID19Regional2023}
{Rojas-Campos}, A., Stelz, L., and Nieters, P. (2023).
\newblock Learning {{COVID-19 Regional Transmission Using Universal
  Differential Equations}} in a {{SIR}} model.
\newblock \doi{10.48550/ARXIV.2310.16804}.

\bibitem[{Santana and Costa(2023)}]{santanaEfficientHybridModeling2023}
Santana, V.V. and Costa, E. (2023).
\newblock Efficient hybrid modeling and sorption kinetic model discovery for
  non-linear advection-diffusion-sorption systems: {{A}} systematic scientific
  machine learning approach.

\bibitem[{Steinebach(2023)}]{steinebach}
Steinebach, G. (2023).
\newblock Construction of rosenbrock--wanner method rodas5p and numerical
  benchmarks within the julia differential equations package.
\newblock \emph{BIT Numerical Mathematics}, 63(2), 27.

\bibitem[{Tang(2006)}]{tang}
Tang, K.T. (2006).
\newblock \emph{Mathematical methods for engineers and scientists}, volume~2.
\newblock Springer.

\bibitem[{Telen et~al.(2012)Telen, Logist, Van~Derlinden, Tack, and
  Van~Impe}]{telen}
Telen, D., Logist, F., Van~Derlinden, E., Tack, I., and Van~Impe, J. (2012).
\newblock Optimal experiment design for dynamic bioprocesses: a multi-objective
  approach.
\newblock \emph{Chemical Engineering Science}, 78, 82--97.

\bibitem[{Telen et~al.(2014)Telen, Vercammen, Logist, and Van~Impe}]{telen2}
Telen, D., Vercammen, D., Logist, F., and Van~Impe, J. (2014).
\newblock Robustifying optimal experiment design for nonlinear, dynamic (bio)
  chemical systems.
\newblock \emph{Computers \& Chemical Engineering}, 71, 415--425.

\bibitem[{Uci{\'n}ski and Bogacka(2005)}]{ucinski}
Uci{\'n}ski, D. and Bogacka, B. (2005).
\newblock T-optimum designs for discrimination between two multiresponse
  dynamic models.
\newblock \emph{Journal of the Royal Statistical Society Series B: Statistical
  Methodology}, 67(1), 3--18.

\bibitem[{Van Der~Ploeg et~al.(2014)Van Der~Ploeg, Austin, and
  Steyerberg}]{van}
Van Der~Ploeg, T., Austin, P.C., and Steyerberg, E.W. (2014).
\newblock Modern modelling techniques are data hungry: a simulation study for
  predicting dichotomous endpoints.
\newblock \emph{BMC medical research methodology}, 14, 1--13.

\bibitem[{Versyck et~al.(1997)Versyck, Claes, and Van~Impe}]{versyck}
Versyck, K.J., Claes, J.E., and Van~Impe, J.F. (1997).
\newblock Practical identification of unstructured growth kinetics by
  application of optimal experimental design.
\newblock \emph{Biotechnology progress}, 13(5), 524--531.

\end{thebibliography}

\end{document}